\documentclass[runningheads]{llncs}
\usepackage{graphicx}
\usepackage{color}
\usepackage{array}
\usepackage{amsmath}
\usepackage{mathtools}
\usepackage{hyperref}
\usepackage{caption}
\usepackage{subcaption}
\let\footnote\thanks
\begin{document}
\title{Scene Graph Generation with Geometric Context}
%
%
\author{{Vishal Kumar\thanks{Equal contributors}} \and
Albert Mundu$^\star$ \and
Satish Kumar Singh}
\authorrunning{Mundu et al.}
%
\institute{Indian Institute of Information Technology Allahabad
\\
\small{\email{vishal.rishu26@gmail.com, phc2016001@iiita.ac.in, sk.singh@iiita.ac.in}}}
\maketitle              
\begin{abstract}
Scene Graph Generation has gained much attention in computer vision research with the growing demand in image understanding projects like visual question answering, image captioning, self-driving cars, crowd behavior analysis, activity recognition, and more. 
Scene graph, a visually grounded graphical structure of an image, immensely helps to simplify the image understanding tasks. In this work, we introduced a post-processing algorithm called Geometric Context to understand the visual scenes better geometrically. 
We use this post-processing algorithm to add and refine the geometric relationships between object pairs to a prior model. We exploit this context by calculating the direction and distance between object pairs. We use Knowledge Embedded Routing Network (KERN) as our baseline model, extend the work with our algorithm, and show comparable results on the recent state-of-the-art algorithms.

\end{abstract}
\section{Introduction}
Image contains much information and understanding it has become a challenging task in computer vision. With recent breakthroughs \cite{fast-rcnn,faster-rcnn,wu2019detectron2,detr2020} in object detection, there is a growing demand for scene graph generation as it helps in scene understanding tasks like visual question answering, image captioning, self-driving car and crowd-behavior analysis. Scene graphs are used to represent the visual image in a better and more organized manner that exhibits all the possible relationships between the object pairs. 
The graphical representation of the underlying objects in the image showing relationships between the object pairs is called a scene graph \cite{johnson2015image}.  

Object detection is important but detecting only the objects in an image is insufficient to understand the scene. For example, in an image of a woman with a motorcycle, the woman may be riding the motorcycle, standing beside it, or simply holding it. Here, the object pair is `woman' and `motorcycle'; the possible relationships (predicates) are `riding', `standing', and `holding'. 
Understanding these distinct relationships between the object pairs is important for adequate scene understanding. 
Relational reasoning between different objects or regions of random shape is crucial for most recent notable tasks such as image captioning, self-driving cars (cite recent projects) in the computer vision domain. 

Visual Relationship Detection refers to detecting and localizing the objects present in the image and finding associated relationships $<predicate>$ between the object pairs $<object - subject>$.  The object pairs $<object - subject>$ are related by $<predicate>$ and this relationship is represented as a triplet of $<object - predicate - subject>$ in the scene graph. 
In this work, we first detect and localize the object pairs and classify the interaction or the predicate between each of the object pairs. It is similar to object detection but with a large semantic space in which the possible relationships between each pair are much higher than the objects. Reasoning and analyzing such possible combinations of relationships and organizing them in a graph enhances the understanding of the scene broadly.  

It is observed that in any image, the objects present in it are highly correlated, and the pattern is repeatedly occurring \cite{motifs}. For example,``man" and ``woman" both wear clothes, ``car" tends to have ``wheels"; such strong regularities can be found between the different objects. Thus, the distribution of real-world relations becomes highly skewed and unbalanced.  The existing models perform better when the relationships are more frequent and poorly when the relationships are less frequent. \cite{kern} introduces a structured knowledge graph that includes the correlations among the existing objects and the relationships between the pairs to address this unbalanced distribution problem. However, the spatial geometric relationships, which are highly present in the real-world distribution, significantly contribute to understanding the scene and are less exploited in the existing works.  The relationships such as near, far, top, down, beside, left, right are geometric relationships that frequently appear. Our work aims to add geometric relationships and refine the relations that the baseline model predicts.

\section{Related Work}
In recent computer vision researches, detection algorithms have made tremendous breakthroughs in detecting and localizing the objects in an image. With such achievements,  the research has gained much attention to visual relationship detection. Detecting only the objects in the image does not help in understanding the underlying scene. However, detecting objects and the relationships between the object pairs can lead to proper understanding and improve detection based on the scene's context. 

Initial works \cite{vrd,imp,motifs,kern} have used the object detectors \cite{fast-rcnn,faster-rcnn} to extract the region proposals of the objects and tried to improve scene graph generation task by incorporating 1) \textbf{language priors} \cite{vrd} from semantic word embeddings, 2) \textbf{co-occurrence matrix} \cite{kern} of objects and their relationships and 3) \textbf{motifs} \cite{motifs}. Due to sparse distribution of relationships in the existing datasets \cite{vrd,visualgenome}, Chen et al. \cite{limitedlabels} tried to generate scene graphs from limited labels using few-shot learning. Also, Tang et al. \cite{Tang_2020_CVPR} observed that the scene graph generation task performs poorly due the severe training bias and thus introduced scene graph generation (SGG) framework based on the casual inference. 
\subsection{\textbf{Visual Relationship Detection}}
The main objective of this model was to detect the relationships between object pairs, i.e., visual relationships in an image. 
\cite{vrd} proposed two modules: 1) \textbf{visual appearance module}, which learns the occurrence of predicate and object relationships. After learning the appearance, it merges them to anticipate the visual relationships between objects jointly.
2) \textbf{language module} helps to recognize the closer predicates with the help of pretrained word vectors. It converts visual connections into a vector space, from which the module optimizes them depending on their proximity to one another. It projected all the relationships into a feature space, allowing semantically related relationships to be closer together. 
\cite{vrd} released a full-fledged dataset named VRD that is now standard for state-of-the-art comparisons. The dataset contains images with labeled objects and annotations from which the relationships between the object pairs are inferred. Following \cite{vrd}, the several works \cite{imp,kern,motifs,Tang_2020_CVPR,graph} have addressed unique problems in the SGG task and solved them by introducing their own techniques with major improvements.

\subsection{\textbf{Graph Inference}}
\cite{imp} proposed the first end-to-end model that generates scene graphs by inferring the visual relationships across every object pair of an image.  The model uses standard RNNs to reason the visual relationships between object pairs and uses message passing iteratively to improve the predictions by passing contextual cues between the nodes and edges of the model architecture. \cite{motifs} proposed a model that predicts the most frequent relations among the object pairs; uses alternating highway-LSTMs to address the graph inference problem to detect the objects' and predicates' labels using global context. \cite{kern} addresses the unbalanced distribution issues of the relationships in the existing datasets \cite{vrd,visualgenome} and uses the statistical correlation knowledge graph to regularize the semantic space between object pairs and their relationships. \cite{kern} uses Graph Gated Neural Network to route the knowledge graph through the network iteratively and predicts the labels and relationships of the objects. \cite{Tang_2020_CVPR} observed that the scene graph generation task could perform poorly due to training bias. Hence, based on the causal inference, they proposed SGG framework, which would draw counterfactual causality from the trained graph to infer the effect from bad bias, which should be removed. In this work, we use KERN's routing network \cite{kern} to infer the objects and their relationships.

\subsection{\textbf{Routing Network with Embedded Knowledge}}

Chen et al. \cite{kern} proposed a network that is based on knowledge routing, which learns the objects and relationships between them by embedding a statistical correlation knowledge graph. The network uses Faster RCNN \cite{faster-rcnn} to detect a collection of region proposals in an image. A graph is created to connect the regions, and a graph neural network is utilized to generate contextually relevant descriptions to determine the class label for every region. For every object pair with the predicted label, we use a graph neural network to predict their predicates, and we create an additional graph to link the provided pair of objects, including all possible predicates. The scene graph is generated once the process is completed for all object pairs. We employ this routing network in our work and extend it with a geometric context post-processing algorithm.

\section{\textbf{Proposed Solution}}
To have more insights on the dataset \cite{visualgenome}, we categorized the relations based on higher relation types - \textbf{Geometric}, \textbf{Possessive}, \textbf{Semantic}, \textbf{Misc.}, similar to \cite{motifs}. In Table \ref{table:relation-types}, geometric and possessive types dominate the whole dataset. However, the geometric relations that are crucial to scene understanding in indoor and outdoor scenes are yet to be exploited. In this work, we estimate the geometric relationships of every object pairs using geometric parameters - \textbf{distance}, \textbf{direction} shown in Figure \ref{fig:gcontext}, after the region proposals are regenerated and validate them on the model's output \cite{kern}. We append the geometric relationships if the reasoning module of the model fails to predict; also, we filter the geometric relations if the model's predicted relations become too ambiguous. As the dataset \cite{visualgenome} has 15 geometric relations, we further categorized them based on the parameters. For example, if the \textbf{distance} between the pair of the objects is small, we categorize it as \textbf{near} and \textbf{far} on the other hand. Moreover, estimating the \textbf{direction}, we further categorize the geometric relations as top, bottom, under, left, right.

\subsection{Geometric Context}
\begin{figure}
    \centering
    \includegraphics[width=0.7\textwidth]{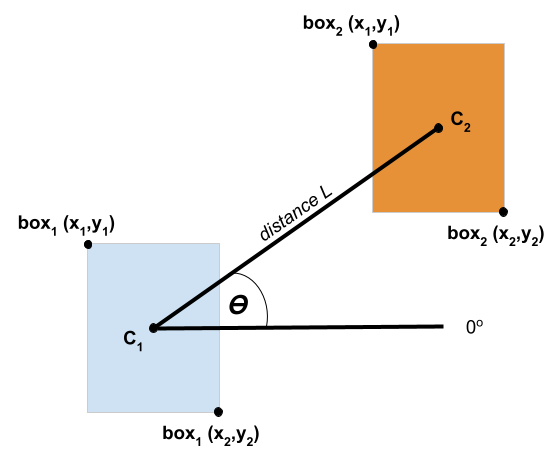}
    \caption{Calculation of Geometric Parameters: \textit{distance} $L$ and \textit{angle} $\theta$. We use $L$ to estimate the nearness of the objects. $\theta$ is used to estimate the direction of the objects with respect to every object. $C_i$ where $i \in \{1,2\}$ is the centroid of $box_i$; $(x_i,y_i)$ is the coordinate of the $box_i$}
    \label{fig:gcontext}
\end{figure}

Using the baseline model \cite{kern}, we extract the model's predicted object labels $(o_i)$, relations $(r_{i\rightarrow j})$, bounding boxes $(B_i)$ and triplets $<o_i,r_{i\rightarrow j},o_j>$ for post-processing. 
We take the bounding boxes $(B_i)$ as input to the proposed algorithm, and calculate the two parameters - 1) \textbf{distance $L$} and 2) \textbf{direction $\theta$}. For calculating $L$ and $\theta$, we first find the centroids of the boxes $B_i$ and $B_j$ as $C_i$, $C_j$ respectively. Taking center coordinates, we use \textbf{L2 Distance} to calculate the distance $L$ between the object pairs; trigonometric function to calculate the direction $\theta$ from $o_i$ to $o_j$ as illustrated in Figure \ref{fig:gcontext}. We perform this operation for all possible object pairs detected by the baseline model.

\begin{figure}%
    \centering
    \includegraphics[width=.8\textwidth]{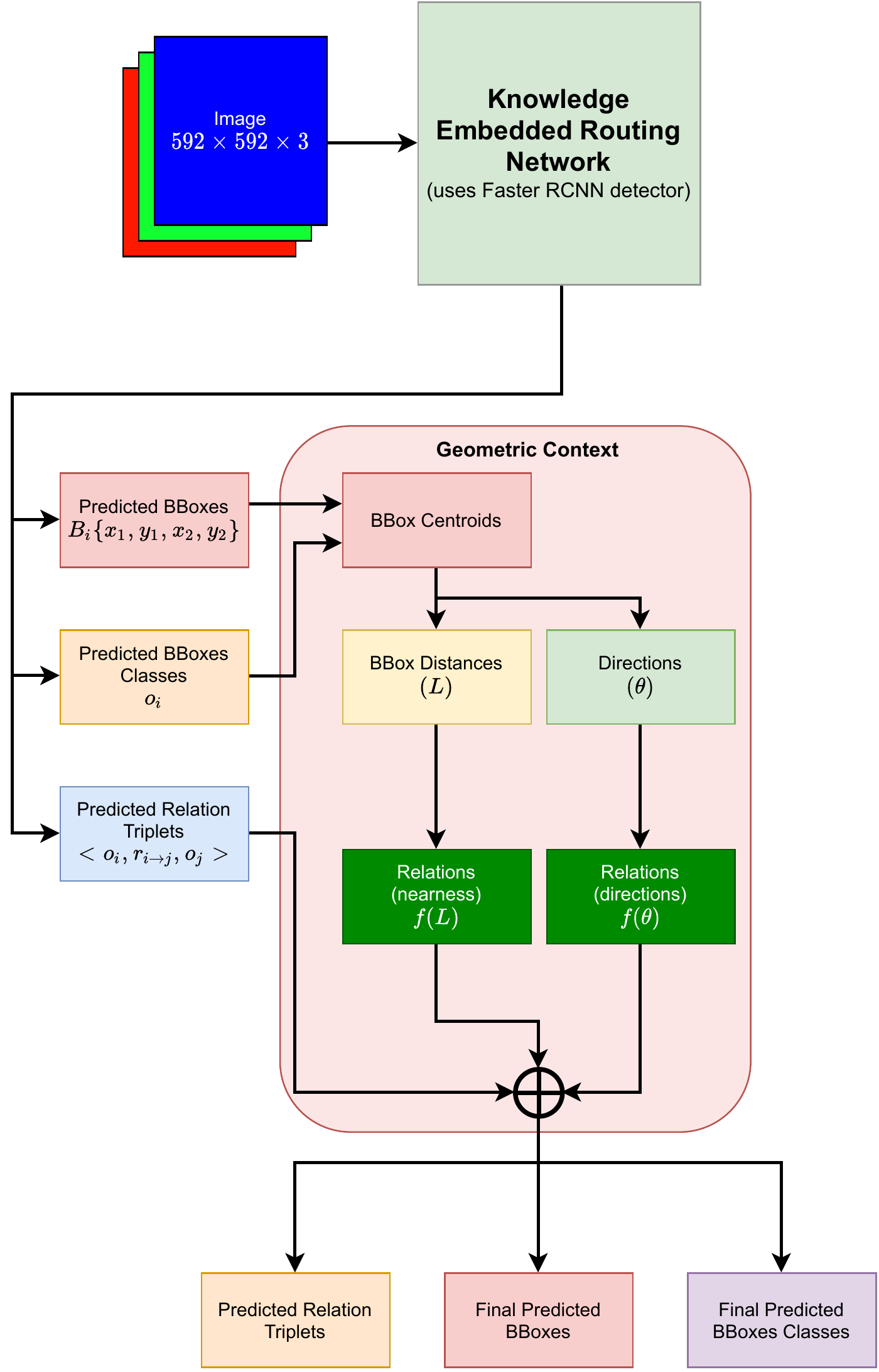}
    \caption{Proposed architecture: \textbf{Geometric Context}. We feed an input image to the KERN \cite{kern} and extract the predicted bounding boxes ($B_i$), classes ($o_i$), and relation triplets $<o_i,r_{i\rightarrow j},o_j>$ from the model.  Using $B_i$ and $o_i$, we calculate the centroids of the bounding boxes. We use the centroids to calculate $L$ and $\theta$ as illustrated in Figure \ref{fig:gcontext}. $(i,j) \in \{1,\dots,N\}$, $N$ is the number of predicted bounding boxes. $\oplus$ denotes concatenation of the predicted relations, addition of bounding boxes and classes (if any) into three categories similar to KERN's output.}
    \label{fig:method}
\end{figure}

Based on these parameters, how do we categorize the geometric relations defined in the dataset? We decide on categorizing the relations based on the following functions - 

\begin{equation}
\label{eqn:1}
    f(\theta)= 
\begin{cases}
    r_1,& \text{if} \; -45^{\circ} < \theta \leq 45^{\circ}\\
    r_2,& \text{if} \;  -135^{\circ} < \theta \leq -45^{\circ}\\
    r_3,& \text{if}  \; 45^{\circ} < \theta \leq 135^{\circ}\\
    r_4,& \text{if}   \; \theta > 135^{\circ} \; \text{or} \; \theta \leq -135^{\circ}
\end{cases}
\end{equation}

where $r_i$ is the relation, $i \in [1, 4]$ and $r_i$ represents `right', `top', `left', and `down' respectively.
\begin{equation}
\label{eqn:2}
    f(L)= 
\begin{cases}
    l_1,& \text{if} \; L < \sqrt(l_{box}^2 + h_{box}^2)/2
    \\
    l_2,& \text{else}
\end{cases}
\end{equation}

where $l_1$ and $l_2$ represents predicates `near' and `far' respectively; \textit{$l_{box}$}, \textit{$h_{box}$} denotes length and height of the bounding box. We concatenate the results of equations \ref{eqn:1} and \ref{eqn:2} with the baseline's model \cite{kern} triplets to add and refine the geometric relations as shown in Figure \ref{fig:method}.

In the result section \ref{sec:results}, we show how the relationships between objects are predicted accurately after the post-processing algorithm. In short, we extract the predicted bounding boxes of an image and the predicted classes using KERN model. After calculating the centroids of the bounding boxes and the distance between them. We then calculate the directions between each bounding boxes. Using the distance and direction, we infer geometric relationships between the objects.
We also added 6 predicate classes to our visual genome dataset \cite{visualgenome}. Two predicate classes - above and near was already present in the state-of-the art dataset. 

\begin{figure}
    \centering
    \includegraphics[width=\textwidth]{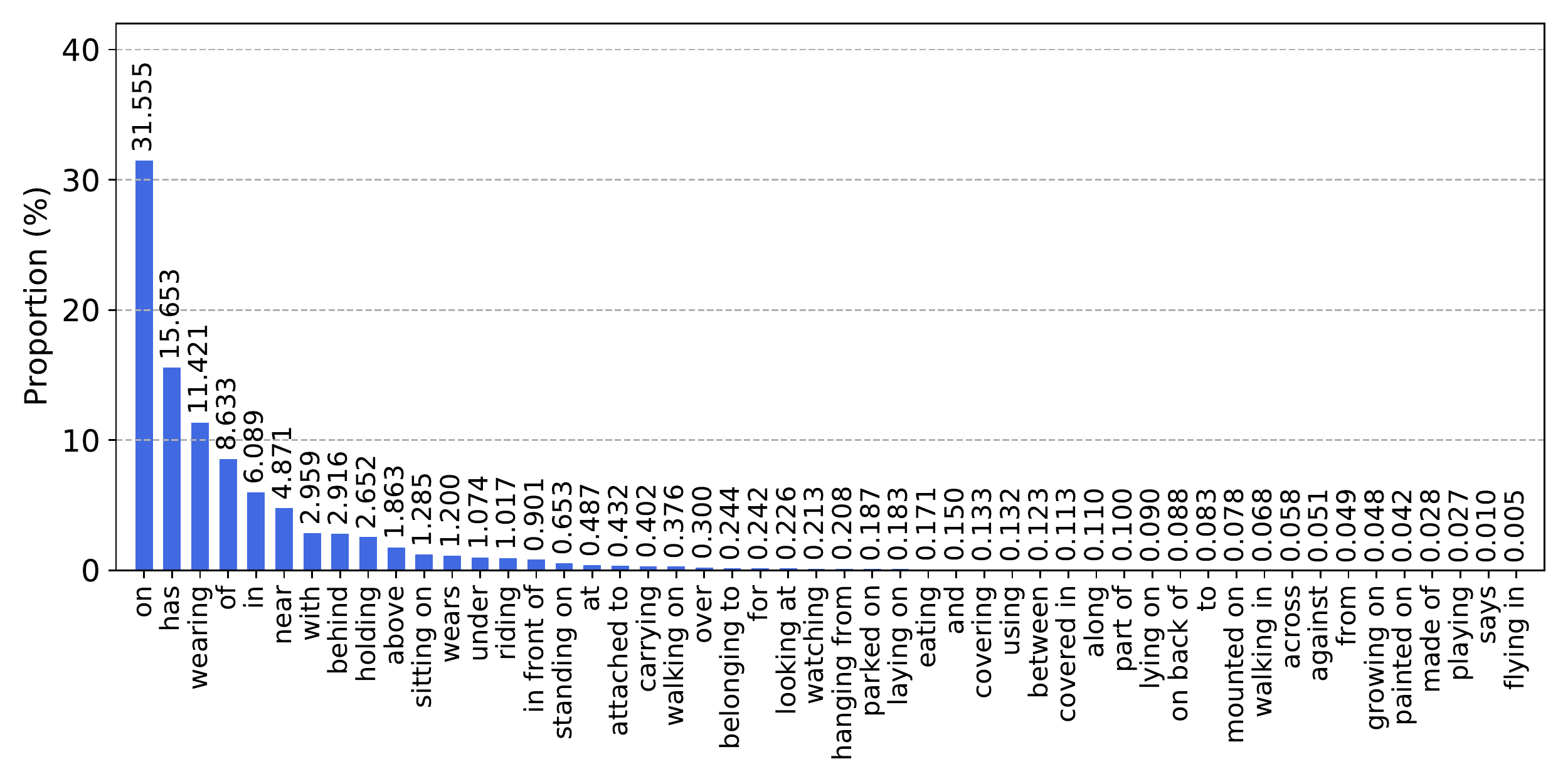}
    \caption{Frequency distribution of relations (predicates) in Visual Genome dataset \cite{visualgenome}}.
    \label{fig:freq}
\end{figure}

\section{\textbf{Dataset}}
\subsection{\textbf{Visual Genome}}
The visual genome dataset \cite{visualgenome} contains 1,08,077 images, 5.4M region descriptions, 1.7M visual question answers, 3.8M object instances, 2.8M attributes and 2.3M relationships. It has 150 object classes and 50 unique relationships with the frequency histogram shown in Figure \ref{fig:freq}. Table \ref{table:relation-types} shows higher relation types of the dataset. The dataset is used for scene graph generation, scene understanding, image retrieval, image captioning, and visual question answering.

\begin{table}[!ht]
\caption{Types of relations in Visual Genome dataset \cite{motifs}. 50\% of the relations in the dataset are geometric relations, followed by 40\% possessive relations. }
\begin{center}
\begin{tabular}[width=\textwidth]{|m{2cm}|p{5cm}|c|c|}
\hline
 \textbf{Types}     &       \textbf{Examples}       &       \textbf{\#Classes}       &   \textbf{\#Instances}\\\hline\hline
 Geometric      &       near,far,under          &       15          &       228k (50.0\%)   \\\hline
 Possessive     &       in,with                 &       8           &       186k (40.9\%)   \\\hline
 Semantic       &      eating,watching,riding   &       24          &       39k (8.7\%)     \\\hline
 Misc.          &       made of,from,for        &       3           &       2k (0.3\%)      \\\hline
\end{tabular}
\end{center}

\label{table:relation-types}
\end{table}

\section{\textbf{Implementation}}
We use Faster RCNN\cite{faster-rcnn} detector to construct the candidate region set, similar to previous efforts \cite{imp,motifs} for scene graph construction. As in \cite{imp,motifs}, the detector uses VGG16-ConvNet\cite{vgg16} as its backbone network, which has been pretrained on ImageNet\cite{imagenet}. Both GGNN's iteration step is set at three. We employ anchor sizes and aspect ratios similar to YOLO-9000\cite{yolo} and initialize input image size to 592$\times$592, as described in \cite{motifs}. We then utilize the stochastic gradient descent(SGD) technique with the strength of a moving object as 0.9, the number of training examples utilized in one iteration equals 2 and 0.0001 weight loss to train the model that detects the regions on the target dataset.
The learning rate for our model is initialized at 0.001 and then divided by ten whenever the validation set's mean average precision reaches a plateau. After that, we use the Adam method and set the number of training examples utilized in one iteration equals two and the strength of a moving object as 0.9, 0.999 to train our entirely-associated layers in addition to the neural network with layered graphs, freezing the parameters of all the present convolution layers in this activity. We set the initial tuning parameter as 0.00001, then split it by ten when the validation set fraction of the relevant sets that are successfully retrieved reaches a plateau.

\subsection{\textbf{Tasks}}
The goal of a scene graph creation is to predict the collection of triplets of $<object-predicate-subject>$. We are assessing our suggested technique with three tasks, previously seen in \cite{imp}.
\begin{itemize}
    \item \textbf{Predicate classification (PredCls)}: Given the box regions of the object pairs, the task is to predict the predicate.
    \item \textbf{Scene graph classification (SGCls)}: Given the box regions, the task is to predict the object categories of the boxes and the predicate between the pairs of the objects. 
    \item \textbf{Scene graph generation (SGGen)}: This task is to detect the box regions, object categories, and the predicates between each detected object pair.  
\end{itemize}
\subsection{\textbf{Evaluation metrics}}
The recall@K (abbreviated as R@K) measure is used to evaluate all scene graph generation approaches. It measures the proportion of actual relationship triplets in an image's best K predictions of the triplet. The breakdown of various connections in the visual genome dataset is quite unequal, and the performance of the most common relationships easily dominates this statistic. Hence, we also use a new metric, mean recall@K (abbreviated as mR@K), to analyze every predicate effectively. This measure computes R@K for each relationship's samples separately, then mR@K is calculated by averaging R@K among all connections. R@K  computed with the constraint setting \cite{kern} only extracts one relationship.
In order to get multiple relationships, other studies ignore this limitation, resulting in higher recall values\cite{pixelgraph}.
For comparisons, we evaluate the Recall@K and the mean Recall@K with constraints.
\section{\textbf{Results}}
\label{sec:results}
\begin{table}
  \caption{Comparison with state-of-the methods on mean Recall@K.}
  \renewcommand{\arraystretch}{1}
    \resizebox{\columnwidth}{!}{%
    \begin{tabular}{|c|c|c c|c c|c c|c|}
    \hline
    & \textbf{Method} & 
    \multicolumn{2}{c|}{\textbf{SGGen}} &  \multicolumn{2}{c|}{\textbf{SGCls}} &
    \multicolumn{2}{c|}{\textbf{PredCls}} &
    \textbf{Mean}
    \\
    &  &\textbf{mR@50} & \textbf{mR@100} & \textbf{mR@50} & \textbf{mR@100} & \textbf{mR@50} & \textbf{mR@100} & \\
    \hline
    \hline
     & IMP\cite{imp} & 0.6 & 0.9 & 3.1 & 3.8 & 6.1 & 8.0 & 3.8\\
    Constraint & SMN\cite{motifs} & 5.3 &6.1 & 7.1 & 7.6 & 13.3 & 14.4 & 9.0\\
     & KERN\cite{kern} & 6.4 & 7.3 & 9.4 & 10.0 & 17.7 & 19.2 & 11.7\\
     &\textbf{Ours} & 6.4 & 7.3 & \textbf{9.5} & \textbf{10.1} & \textbf{17.9} & \textbf{19.4} & \textbf{11.8}\\
     \hline
  \end{tabular}
  }
\label{table:res-1}
\end{table}

The scene graph generation task is evaluated using Visual Genome \cite{visualgenome}, which is the most significant benchmark for analyzing the problem of scene graph generation. In this section, we show and compare our proposed technique with existing works - visual relationship detection(VRD) \cite{vrd}, iterative message passing (IMP) \cite{imp}, neural motif (SMN) \cite{motifs} and knowledge embedded routing network (KERN) \cite{kern}.
\\

We show the results on these methods for the given three tasks in mR@K and R@K. As we can see in Table \ref{table:res-1}, our algorithm outperforms other methods in the constrained setting. However, it does not give any significant improvements over the other methods in the unconstrained setting since the predicates provided by our algorithm are less significant compared to previous predicates. 
\\
\begin{figure}[ht]
\begin{subfigure}{.5\textwidth}
  \centering
  \includegraphics[width=\linewidth]{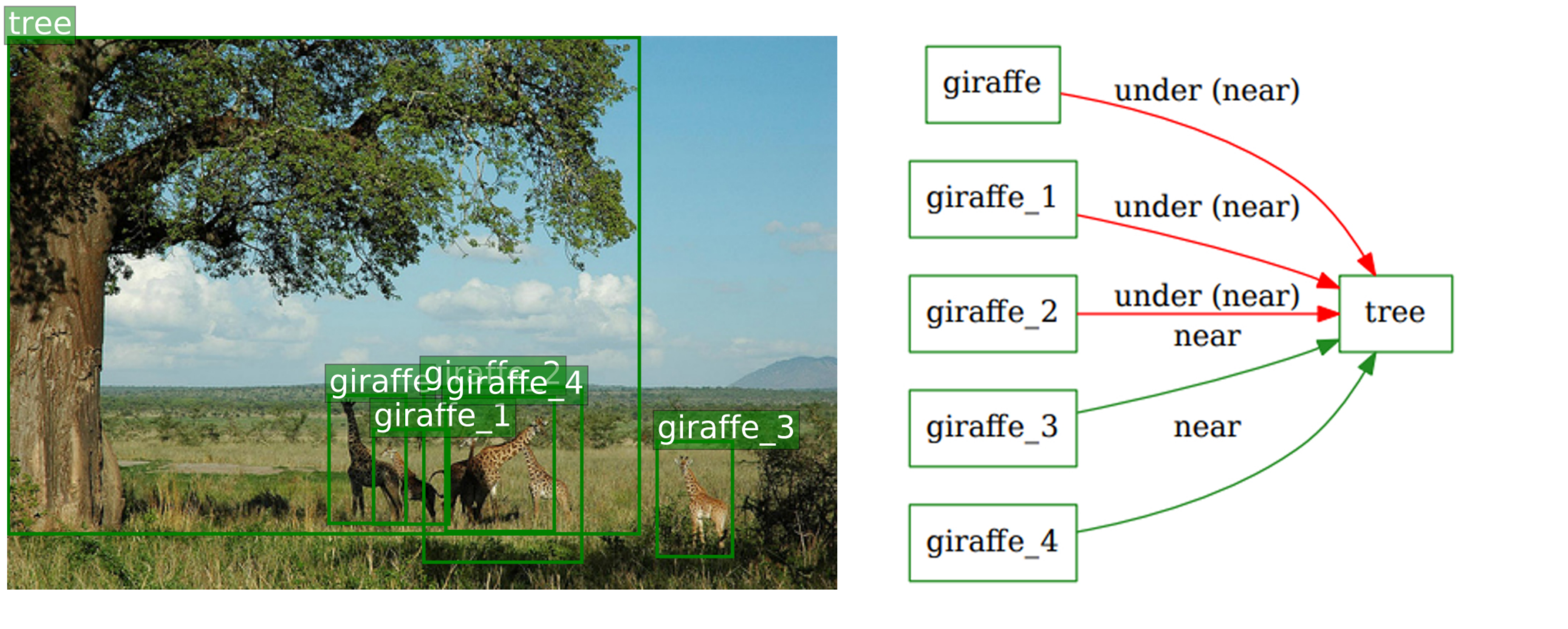}  
\end{subfigure}
\begin{subfigure}{.5\textwidth}
  \centering
  \includegraphics[width=\linewidth]{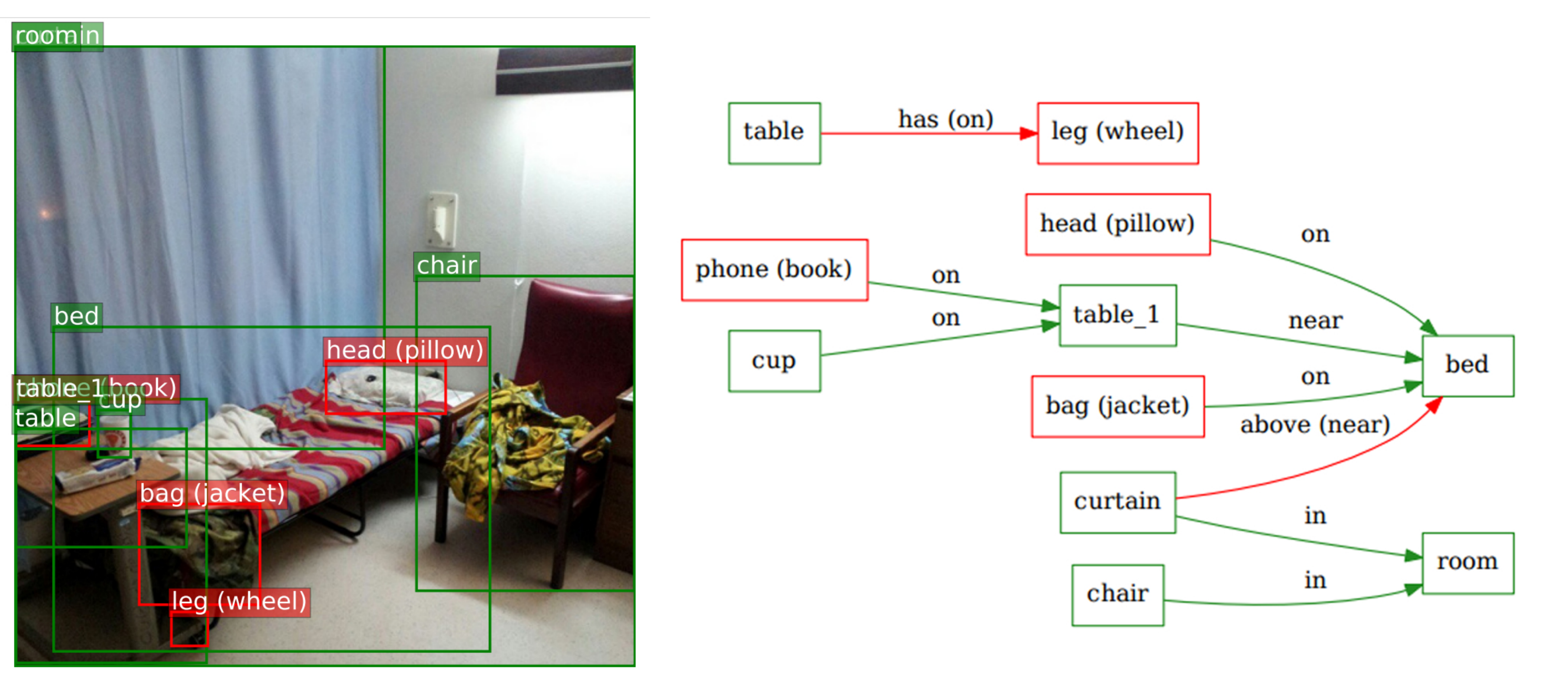}  
\end{subfigure}
\caption{Qualitative results from our model in the Scene Graph Generation (SGGen) setting. The green boxes are the correct predicted boxes and the red boxes are the boxes from the ground truth boxes which the model failed to predict. The green and red edges are the correct predicted relations(predicates) and ground truth relations (model failed to predict) respectively. }
\label{fig:res-img}
\end{figure}

Table \ref{table:res-2} shows the Recall@50 and Recall@100 values on the three major tasks on the Visual Genome dataset for a complete comparison with existing techniques. As we can see in the Table \ref{table:res-1} and \ref{table:res-2}, our method gives good results for both the metrics mR@K and R@K in the constraint setting.
We also compared our work on predicate classification recall values with KERN \cite{kern} results (Table \ref{table:res-3}). As we can see in the results, only two predicates - `above' and `near' has better results than the KERN model. Figure \ref{fig:res-img} shows the qualitative results produced by the model after the geometric context algorithm.
\begin{table}
 \caption{Comparison with state-of-the methods on Recall@K}
  \label{table:res-2}
  
    \begin{tabular}{|c|c|c c|c c|c c|c|}
   
    \hline
    & \textbf{Method} & 
    \multicolumn{2}{c|}{\textbf{SGGen}} &  \multicolumn{2}{c|}{\textbf{SGCls}} &
    \multicolumn{2}{c|}{\textbf{PredCls}} &
    \textbf{Mean}
    \\
    &  &\textbf{R@50} & \textbf{R@100} & \textbf{R@50} & \textbf{R@100} & \textbf{R@50} & \textbf{R@100} & \\
    \hline
    \hline
    & VRD\cite{vrd} & 0.3 & 0.5 & 11.8 & 14.1 & 27.9 & 35.0 & 14.9 \\
     & IMP\cite{imp} & 3.4 & 4.2 & 21.7 & 24.4 & 44.8 & 53.0 & 25.3\\
    Constraint & SMN\cite{motifs} & \textbf{27.2} & \textbf{30.3} & 35.8 & 36.5 & 65.2 & 67.1 & 43.7\\
     & KERN\cite{kern} & 27.1 & 29.8 & 36.7& 37.4 & 65.8 & 67.6 & 44.1\\
     &\textbf{Ours} & 27.1 & 29.8 & \textbf{36.9} & \textbf{37.8} & \textbf{66.1} & \textbf{68.1} & \textbf{44.3}\\
     \hline
  \end{tabular}

\end{table}
\begin{table}
  \centering
   \caption{Compared our work with KERN model on predicate classification recall@50 and recall@100.}
       \label{table:res-3}
  \renewcommand{\arraystretch}{1}
  
    \resizebox{7cm}{!}{%
    \begin{tabular}{|c|c c|c c|}
    \hline
    \textbf{Predicate} & 
    \multicolumn{2}{c|}{\textbf{KERN}} &  \multicolumn{2}{c|}{\textbf{Ours}} 
    \\
      &\textbf{R@50} & \textbf{R@100} & \textbf{R@50} & \textbf{R@100} \\
    \hline
    \hline
    above & 17.0 & 19.4 & \textbf{20.4} & \textbf{25.6}\\
    near & 38.8 & 45.5 & \textbf{42.0} & \textbf{51.1}\\
    at & 32.2 & 37.3 & 32.2 & 37.3\\
    has & 78.8 & 81.3 & 78.8 & 81.3\\
    wearing & 95.8 & 97.1 & 95.8 & 97.1\\
    \hline
  \end{tabular}
    }

\end{table}
\newpage

\section{\textbf{Conclusion}}
Using only the object co-occurrence knowledge of an image is still insufficient to predict the relationship between object pairs effectively. If we provide geometric context with object co-occurrence knowledge, we can further improve the relationship predictions. In this work, we use geometric context with the object co-occurrence knowledge \cite{kern} and achieved relatively better results than existing methods. The future scope of this algorithm would be to evaluate its capability in different scene graph models and check its significance. 
%
%
%

\pagebreak
\bibliographystyle{splncs04}
\bibliography{ref}

\end{document}